\documentclass{article}

\usepackage[main,final]{neurips_2026}

\usepackage[utf8]{inputenc} 
\usepackage[T1]{fontenc}    
\usepackage{hyperref}       
\usepackage{url}            
\usepackage{booktabs}       
\usepackage{amsfonts}       
\usepackage{nicefrac}       
\usepackage{microtype}      
\usepackage{xcolor}         

\usepackage{graphicx}

\usepackage{pifont}
\usepackage{threeparttable} 
\usepackage{multirow}
\usepackage{subfigure}
\usepackage{amsmath}
\usepackage{amssymb}
\usepackage{enumitem}
\definecolor{darkgreen}{rgb}{0.0, 0.5, 0.0}

\usepackage[fixed]{fontawesome5}
\usepackage{pythonhighlight}
\usepackage{caption}
\usepackage{wrapfig}
\usepackage{makecell}  

\definecolor{posgreen}{RGB}{0, 150, 0}
\definecolor{negred}{RGB}{220, 0, 0}
\usepackage{arydshln}

\newcommand{\up}[1]{_{\textcolor{posgreen}{\uparrow #1}}}
\newcommand{\downv}[1]{_{\textcolor{negred}{\downarrow #1}}}

\title{UniDial-EvalKit:
A Unified Toolkit for Evaluating Multi-Faceted Conversational Abilities}

\author{Qi Jia$^1$ \quad  Haodong Zhao$^{2}$ \quad  Dun Pei$^1$ \quad  Xiujie Song$^2$ \quad Ye Shen$^{1,2}$ \quad \textbf{Shibo Wang}$^{1}$  \\ \quad \textbf{Zijian Chen$^{1,2}$} \quad \textbf{Zicheng Zhang$^{1}$} \quad \textbf{Xiangyang Zhu}$^1$ \quad \textbf{Guangtao Zhai}$^{1,2}$
    \\[5pt]
	$^1$Shanghai Artificial Intelligence Laboratory \quad $^2$Shanghai Jiao Tong University\\[5pt]
 {{\small \faGithub} \texttt{\href{https://github.com/UniDial/UniDial-EvalKit}{\textcolor{blue}{https://github.com/UniDial/UniDial-EvalKit}}}}
}

\begin{document}

\maketitle

\begin{abstract}
Benchmarking large language models (LLMs) and agents in multi-turn interactive scenarios is essential for understanding their practical capabilities. However, existing evaluation protocols are highly heterogeneous, differing significantly in dataset formats, model interfaces, and evaluation pipelines, which severely impedes systematic comparison. In this work, we present \textbf{UniDial-EvalKit (UDE)}, a unified evaluation toolkit for assessing interactive AI systems. The core contribution of UDE lies in its holistic unification: it standardizes heterogeneous data formats into a universal schema, streamlines complex evaluation pipelines through a modular architecture, and aligns metric calculations under a hierarchical scoring aggregation. It also supports efficient large-scale evaluation through parallel generation and scoring, as well as checkpoint resume to eliminate redundant computation.
Leveraging UDE, we conduct an extensive evaluation across diverse multi-dimensional benchmarks. Our empirical analysis shows that no single system consistently outperforms others across all benchmarks, while current memory agents often fail to surpass full-context baselines. Further analyses highlight several future directions, including benchmark deduplication and more adaptive memory architectures.

\end{abstract}

\section{Introduction}

In the rapidly evolving landscape of large language models (LLMs) and agent-based systems, assessing performance in multi-turn dialogues has become essential for understanding their practical capabilities. However, this remains a significant challenge, largely due to the absence of unified standards and frameworks for dialogue evaluation. Such fragmentation complicates model comparison and limits the ability to track progress across different tasks and scenarios.

While single-turn evaluation frameworks are well-established, such as OpenCompass~\citep{2023opencompass}, VLMEvalKit~\citep{duan2024vlmevalkit} and LMMs-Eval~\citep{zhang2024lmmsevalrealitycheckevaluation}, the unification of multi-turn evaluations remains largely underexplored. Unlike single-turn question answering, evaluating continuous multi-turn interactions introduces unique challenges:

First, multi-turn evaluation \textit{lacks a unified schema} due to inconsistent assessment designs. Existing benchmarks evaluate different interaction scopes, such as only the final response~\citep{deshpande-etal-2025-multichallenge,wu2024longmemeval, hu2025evaluating}, every dialogue turn~\citep{bai2024mt,he2024multi,yan-etal-2025-codeif}, or specific turns~\citep{kwan2024mt,jiayang2026amemgym}. This inconsistency hinders the development of unified evaluation frameworks across benchmarks.

Second, inconsistencies in evaluators and scoring metrics further fragment the evaluation landscape. Existing benchmarks adopt automated evaluators such as LLMs-as-judges~\citep{zheng2023judging,li-etal-2025-generation}, but rely on heterogeneous judge models and scoring frameworks. For example, the benchmark of \citet{fan2025fairmtbench} uses GPT-4 as the judge, while \citet{he2024multi} adopts GPT-4o and Claude. The \textit{lack of unified workflows} hinders fair comparisons and flexible metric adoption.

Third, \textit{the lack of standardized score aggregation} further fragments multi-turn evaluation. Benchmarks employ diverse strategies to summarize turn-level metrics, such as mean~\citep{he2024multi} and minimum~\citep{bai2024mt,cao2026safedialbench} across turns, causing reported scores to partly reflect aggregation choices rather than true model capabilities. This may cause perceived performance differences to stem from aggregation choices rather than genuine capability improvements.

To address these challenges, we introduce \textbf{UniDial-EvalKit (UDE)}, a unified evaluation framework for continuous multi-turn conversational systems, including both LLMs and memory agents. UDE standardizes the evaluation paradigm through three key components. First, it introduces a \textit{unified dialogue schema} to normalize heterogeneous benchmarks into a standardized format. Second, building upon this unified data foundation, UDE implements a \textit{standardized evaluation pipeline} that decouples dataset management, model interaction, and metric evaluation into modular components, enabling flexible integration of new models and metrics. The framework also supports efficient large-scale evaluation through parallel processing and checkpoint resume. Finally, UDE adopts a \textit{hierarchical score aggregation} strategy that systematically aggregates metrics from turn to dialogue and benchmark levels, supporting diverse aggregation methods such as min, mean, and max pooling for unified cross-dialogue evaluation.

By integrating benchmarks across multiple dimensions, UDE \textbf{provides a holistic understanding of model capabilities}. We conduct extensive empirical evaluations across memory, safety, and instruction-following tasks, systematically assessing over ten contemporary LLMs alongside four representative memory agents. Our analysis reveals several notable findings. No single LLM or memory agent achieves universal dominance across benchmarks, reflecting the heterogeneous nature of multi-turn conversational abilities. While thinking-mode models generally obtain stronger overall performance, they may degrade on tasks such as coding and instruction following while incurring longer generation latency. Moreover, augmenting LLMs with external memory agents does not reliably improve long-history interaction capabilities. Existing memory agents remain heavily shaped by person- or event-centric inductive biases, limiting their generalizability to diverse scenarios.

Guided by these findings, we further discuss several \textbf{directions for future research}, including the construction of more scalable and diagnostically informative multi-turn benchmarks with reduced redundancy and finer-grained annotations, as well as the development of adaptive memory mechanisms that better complement increasingly capable long-context LLMs by continuously learning from interactions and adapting to evolving user intentions and preferences. We hope UDE can facilitate more transparent, reproducible, and comprehensive evaluation of multi-turn conversational systems.

\section{UniDial-EvalKit Design}

\begin{figure}
    \centering
    \includegraphics[width=1.0\linewidth]{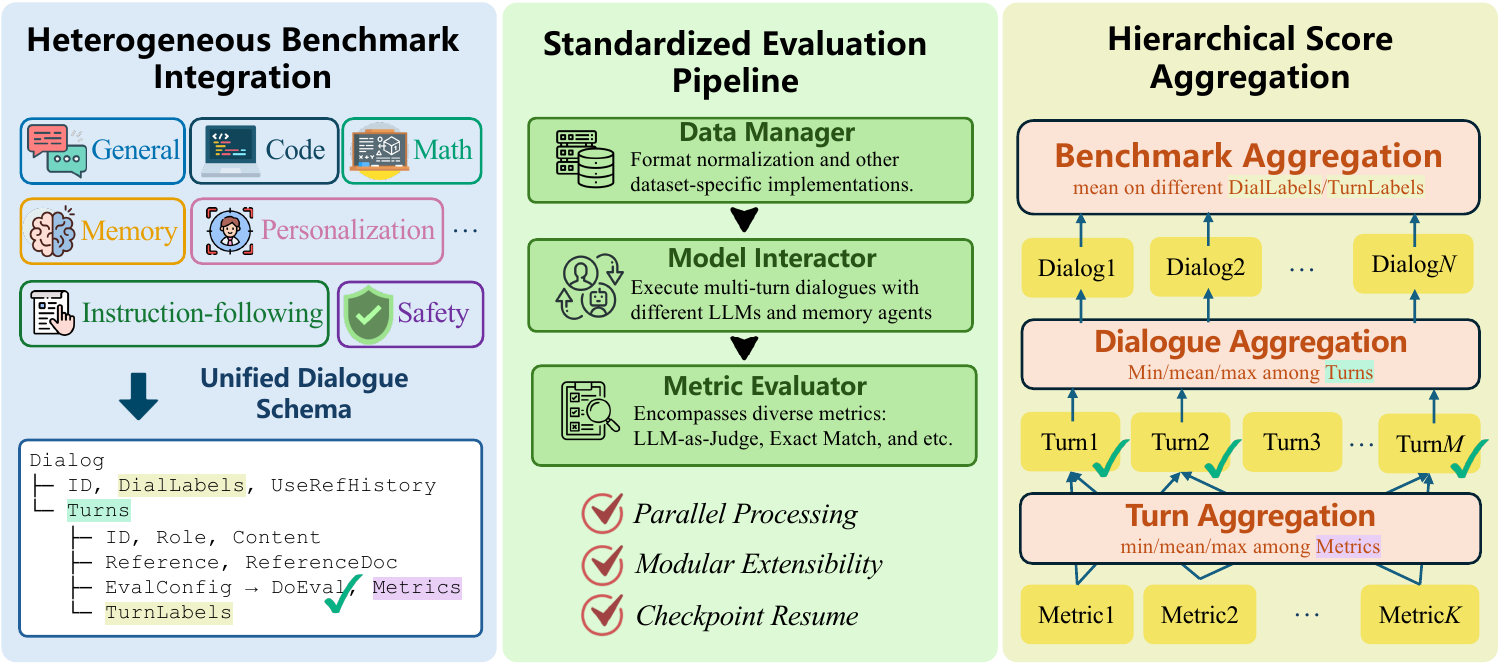}
    \caption{Overview of UniDial-EvalKit.}
    \label{fig:overview}
\end{figure}

UniDial-EvalKit (UDE) is a modular and unified framework for evaluating the multi-turn conversational capabilities of both LLMs and agent systems. UDE achieves normalization across three key dimensions as shown in Figure~\ref{fig:overview}: heterogeneous data sources are unified via a standardized dialogue schema, evaluation processes are standardized through a consistent pipeline, and performance metrics are aggregated according to a unified hierarchical strategy. Internally, the framework implements these principles through four sequential phases: data loading (\textit{DataPhase}), model generation (\textit{GenerationPhase}), metric evaluation (\textit{EvaluationPhase}), and result aggregation (\textit{AggregationPhase}). This highly modular design allows researchers to seamlessly integrate new datasets, custom models or agents, and specialized evaluation metrics without modifying the core pipeline. UDE supports flexible deployment via both a command-line interface (CLI) and a comprehensive Python API (\texttt{EvalPipeline}), ensuring reproducibility and adaptability across diverse experimental setups.\looseness=-1

\subsection{Unified Dialogue Schema}

To harmonize diverse multi-turn dialogue benchmarks, UDE introduces a standardized data schema anchored by the \texttt{Dialog} object. A schematic example is shown in Figure~\ref{fig:overview}, with details in Appendix~\ref{app:schema}. Each instance encapsulates session-level metadata and a chronological sequence of dialogue turns, where each turn records its role, content, optional references, and evaluation configuration.

This schema serves two complementary objectives. First, it enables \textbf{flexible evaluation configurations}. At the dialogue level, \texttt{UseRefHistory} controls how subsequent turns are generated: on-policy generation conditions on model responses, while off-policy generation conditions on ground-truth references. At the turn level, \texttt{EvalConfig} determines whether a turn should be evaluated and specifies the corresponding metrics. Second, the schema preserves hierarchical metadata for \textbf{fine-grained analysis}, including \texttt{DialogLabels}, \texttt{TurnLabels}, and \texttt{ReferenceDoc} for metric aggregation and reference tracking. By decoupling evaluation configurations from label semantics, the unified schema abstracts away dataset-specific idiosyncrasies and enables task-agnostic generation and evaluation.

\subsection{Standardized Evaluation Pipeline}

\paragraph{Data Manager}
The Data Manager handles dataset normalization, where all dialogues are transformed into the standardized \texttt{Dialog} schema to ensure consistency across benchmarks. 

\paragraph{Model Interactor}

UDE includes both base models and memory-enabled agents. Base models~\citep{gpt54, gemini31pro} receive the full dialogue history as input, which tests their internal reasoning and comprehension capabilities. When the dialogue history exceeds the model's supported context length, it is typically truncated~\citep{tan2025prospect, choi2026dycp}. Instead, memory agents~\citep{jimenez2024hipporag, fang2026lightmem} rely on external memory mechanisms to manage history. This mechanism retrieves relevant information~\citep{zhang2025dh, cheng2023lift}, allowing LLMs to perform single-turn question answering based on the curated context.

For memory agents, we implemented standardized input-output functions based on their original repositories to ensure fair comparison of memory capabilities. On the input side, each turn, consisting of a user query and the corresponding assistant response, is treated as a basic unit, which reflects the common setup. On the output side, returned memory entries are processed using a consistent prompt template for model reasoning, and are arranged sequentially to preserve their original priority. This procedure avoids unfairness that may arise from differences in the internal prompts of certain agents~\citep{jimenez2024hipporag,kang2025memory}. See more discussions in Appendix~\ref{app:input_output}.

Additionally, the generation pipeline includes explicit lifecycle management (\texttt{begin\_dialog} and \texttt{end\_dialog}) to maintain coherent contextual memory across multi-turn interactions. Internal states and caches are cleared after each dialogue to prevent cross-dialogue contamination.

\paragraph{Metric Evaluator}

The Metric Evaluator dynamically loads and executes metrics specified in each turn's \texttt{eval\_config}. Supported metrics are managed through a centralized registry (\texttt{METRIC\_REGISTRY}), currently covering a wide range of evaluation protocols including LLM-as-a-Judge~\citep{bai2024mt}, Exact Match~\citep{liang2024mathchat}, Instruction Adherence Accuracy~\citep{he2024multi}, and other task-specific metrics~\citep{rakotonirina-etal-2025-tools}. All metrics are normalized to $[0,1]$ to ensure comparability across different evaluation protocols. 

\begin{table}[h]
    \centering
    \small
    \caption{
    Dataset statistics in UniDial-EvalKit. Columns include the total number of dialogues (\#Dialog), average number of turns per dialogue (\#Turn), average number of words per dialogue (\#Word), whether multiple metrics are applied (Multi-metric), whether generated responses are fed back as dialogue context (On-policy), whether fine-grained grounding labels are provided (Ref. Doc.), whether timestamps are available (Timestamp), and the target conversational {capability} (Capability).
    }
    \begin{tabular}{c|cccccccc}
    \toprule[1pt]
         & \#Dialog & \#Turn & \#Word & \makecell{Multi-\\metric} & \makecell{On-\\policy} & \makecell{Ref.\\Doc.} & \makecell{Time\\stamp} & Capability  \\
    \midrule[1pt]
        LoCoMo & 1,986 & 616.09 & 15,597.69 & \checkmark & \texttimes  & \checkmark & \checkmark & Memory \\
        MathChat & 5,276 & 4.50 & 139.52 & \checkmark & \texttimes   & \texttimes & \texttimes & Math \\
        MemoryCode & 1,512 & 290.06 & 9,047.20 & \texttimes & \texttimes  & \texttimes & \texttimes & Code \\
        MT-Bench-101 & 1,388 & 6.06 & 60.46 & \texttimes & \texttimes  & \texttimes & \texttimes & General \\
        PersonaMem & 589 & 164.90 & 21,549.07 & \texttimes & \texttimes  & \texttimes & \texttimes & Personalization \\
        Multi-IF & 4,501 & 5.98 & 56.83 & \texttimes & \checkmark  & \texttimes & \texttimes & Instr. Follow. \\
        SafeDialBench & 4,053 & 9.53 & 32.20 & \texttimes & \texttimes  & \texttimes & \texttimes & Safety \\
    \bottomrule[1pt]
    \end{tabular}
    \label{tab:data_statistics}
\end{table} 

\subsection{Hierarchical Score Aggregation}
\label{sec:hierarchical_score_aggregation}

UDE aggregates turn-level metrics into session- and dataset-level summaries hierarchically:

\begin{itemize}[leftmargin=0.5cm, noitemsep, topsep=0pt]
\item \textbf{Turn-level:} Each turn may include one or more metrics, for example evaluating multiple instructions for compliance~\citep{he2024multi} or assessing different dimensions of a single response~\citep{cao2026safedialbench}. Scores within a turn are aggregated using configurable pooling methods. 
\item \textbf{Dialogue-level:} Each dialogue contains a variable number of turns. Turn-level summaries are further aggregated to compute a dialogue-level score. By default, minimum aggregation is applied to reflect dialogue-level failures, but alternative strategies can be used depending on whether the focus is on average turn performance or overall interaction experience.
\item \textbf{Dataset-level:} Dialogue-level scores are aggregated across all sessions to produce a final dataset score. Aggregation can be performed by flattening all turns or by averaging dialogue-level summaries, allowing flexibility in capturing either per-turn performance or dialogue-level quality.
\item \textbf{Metric-wise aggregation:} Aggregation can also be performed independently for each metric, allowing fine-grained analysis across distinct performance dimensions.
\end{itemize}

This hierarchical strategy ensures consistent, flexible, and interpretable evaluation results across large-scale, multi-turn dialogue benchmarks. The impact of different choices are in Appendix~\ref{app:aggregation_choice}.

\subsection{Efficiency \& Acceleration}
Given the substantial computational overhead and latency inherent in evaluating multi-turn interactions, UDE is engineered with multithreading and fault-tolerance mechanisms. During both the generation and evaluation phases, the framework leverages \texttt{ThreadPoolExecutor} to enable efficient, dialogue-level concurrent processing. To mitigate network instability or unexpected interruptions, UDE features an automatic resumption capability: it scans the output directory to identify the latest checkpoints and processes only the pending dialogues. This checkpoint-based resumption prevents redundant computation, ensuring stable and efficient large-scale evaluation.

\section{Experimental Setup}
\label{sec:setup}

Based on UDE, we have integrated the multiple benchmarks and evaluation targets, and conduct extensive experiments based on the following evaluation configurations.

\subsection{Benchmarks}
To provide a comprehensive and multifaceted evaluation of multi-turn conversational capabilities, UniDial-EvalKit integrates a diverse suite of benchmarks, including \textbf{LoCoMo}~\citep{maharana2024evaluating}, \textbf{MathChat}~\citep{liang2024mathchat}, \textbf{MemoryCode}~\citep{rakotonirina-etal-2025-tools}, \textbf{MT-Bench-101}~\citep{bai2024mt}, \textbf{PersonaMem}~\citep{jiang2025know}, \textbf{Multi-IF}~\citep{he2024multi}, \textbf{SafeDialBench}~\citep{cao2026safedialbench} and etc. Statistics for normalized benchmarks are shown in Table~\ref{tab:data_statistics} with more details in Appendix~\ref{app:dataset}. Certain datasets were converted from their original formats into a User-Assistant interaction structure, with an injected a system prompt that clarifies participants' roles if necessary.

\subsection{Evaluated Models and Agents}

\paragraph{Models.} Gemma-4-31B~\citep{googledeepminds2026gemma4} and Qwen-3.6-27B~\citep{qwen3.6-27b} are locally deployed on 4 NVIDIA H20 GPUs. The remaining models are accessed via API, with GPT-5.4~\citep{gpt54} and DeepSeek-V3.2~\citep{liu2025deepseek} being a non-thinking mode and Gemini-3.1-Pro~\citep{gemini31pro}, GLM-5~\citep{zeng2026glm}, MiniMax-M2.5~\citep{minimaxm25}, Kimi-K2.5~\citep{team2026kimi}, and Qwen3-Max~\citep{qwen3max} operating in thinking mode.

\paragraph{Agents.} HippoRAG~\citep{jimenez2024hipporag} exemplifies structure-augmented RAG agents. LightMem~\citep{fang2026lightmem} filters and refines input turns to construct topic-coherent memories. RFMem~\citep{zhang2026evoking} either performs direct top-K retrieval or iteratively expands memory evidence based on the defined familiarity. MemPalace~\citep{mempalace} operates in a hybrid mode, extracting different keywords and entities to determine retrieval similarity, followed by reranking memory candidates using an LLM.

\subsection{Evaluation Configuration} 
\label{sec:evaluation_configuration}

To streamline the evaluation process, we first perform stratified sampling based on dialogue labels to reduce the sizes of MemoryCode, MultiIF, and SafeDialBench to 1,000, 1,003, and 1,036 samples, respectively. Pilot experiments detailed in Appendix~\ref{sec:down_sampling} confirm that this down-sampling strategy does not significantly affect model performance. 

During the inference phase, to manage computational costs, each model output is strictly limited to 1,024 tokens per turn, with the exception of MemoryCode, which permits up to 4,096 tokens. This length constraint ensures fair and consistent cross-model comparisons while reflecting real-world efficiency requirements. Models that require excessively long generation or ``thinking'' times can negatively impact user experience, highlighting the critical trade-off between reasoning performance and interactive responsiveness. For memory-augmented agents, we uniformly utilize \texttt{all-MiniLM-L6-v2}~\citep{reimers-2019-sentence-bert} as the embedding model where applicable, configuring the retrieval module to consistently return the top-10 memory snippets.

For metric computation, when utilizing LLM-as-a-judge protocols, GPT-4.1 is adopted as the backbone evaluator following prior work~\citep{li2024crowdsourced}, with the corresponding benchmarks evaluated using their original prompts. Finally, to derive the ultimate results across all systems, we adopt a unified hierarchical aggregation strategy with \texttt{mean-min-dialog} for the three levels in Sec.~\ref{sec:hierarchical_score_aggregation}.

\section{Results}

Based on UDE, we analyze the performance of the tested LLMs and agents.

\subsection{Performance of LLMs}

Results of LLMs are summarized in Table~\ref{tab:llm_performance} with confidence intervals in Appendix~\ref{app:confidence_interval}. Under our evaluation setup, there is no conspicuous performance gap between open-source and proprietary models. Overall, GPT-5.4 achieves the highest average score of 69.97\%, closely followed by the open-weight Gemma-4-31B at 68.98\%. A fine-grained analysis reveals distinct capability profiles. GPT-5.4 demonstrates superior performance in mathematical reasoning and safety alignment, whereas Gemma-4-31B exhibits stronger proficiency in long-context processing and personalized interactions. GLM-5 performs well in memory-intensive information tracking and coding scenarios, while Gemini-3.1-Pro leads in personalization. \textbf{No single model achieves universal dominance across all benchmarks.}

Following~\citet{qian2026benchmark}, we compute \textbf{the discriminability score} of each benchmark to measure its ability to distinguish state-of-the-art models. Empirically, benchmarks with scores above 0.4 provide strong discrimination, while those below 0.2 offer minimal differentiation. We observe that MT-Bench-101 is largely saturated across models, resulting in low discriminability. In contrast, datasets requiring rigorous structural adherence and cumulative state tracking, such as Multi-IF and MemoryCode, exhibit high score variance and strong discriminative power.

\begin{table*}[t!]
    \centering
    \small
    \caption{Performance(\%) of LLMs on normalized benchmarks. \textbf{Bold} indicates the best result, while \underline{underlined} scores denote the second-best performance. Results may differ from original papers due to multi-turn formatting, stricter score aggregation, and distinct evaluation setups.}
    
    \begin{tabular}{l|p{0.8cm}cccccc|c}
        \toprule[1pt]
         & LoCoMo & \makecell{Math\\Chat} & \makecell{Memory\\Code} & \makecell{MT-Bench\\-101} & \makecell{Persona\\Mem} & Multi-IF & \makecell{SafeDial\\Bench} & Avg \\
         \midrule[1pt]
         GPT-5.4 & 61.03 & \textbf{80.80} & \underline{42.26} & 93.23 & \underline{70.12} & \textbf{75.46} & \textbf{66.92} & \textbf{69.97} \\
         Gemma-4-31B & 66.36 & 73.11 & \textbf{44.91} & 92.62 & \textbf{79.29} & \underline{74.95} & 51.64 & \underline{68.98} \\
         Qwen3-Max & 62.11 & 77.31 & 34.06 & \textbf{93.62} & 65.20 & 68.61 & \underline{62.98} & 66.27\\
         Qwen-3.6-27B & 66.80 & 76.81 & 26.39 & \underline{93.43} & 65.37 & 67.21 & 61.29 & 65.33 \\
         DeepSeek-V3.2 & 59.25 & \underline{77.79} & 25.99 & 91.17 & 64.52 & 65.10 & 54.63 & 62.64 \\
         Gemini-3.1-Pro & \underline{69.05} & 75.20 & 39.68 & 89.32 & \textbf{79.29} & 33.98 & 33.79 & 60.04 \\
         GLM-5 & \textbf{69.26} & 69.14 & \textbf{44.91} & 85.47 & 63.16 & 25.88 & 33.40 & 55.89 \\
         MiniMax-M2.5 & 66.04 & 70.54 & 34.13 & 89.98 & 42.78 & 31.79 & 48.75 & 54.86 \\
         Kimi-K2.5 & 42.91 & 60.47 & 31.88 & 90.77 & 55.35 & 26.18 & 51.35 & 51.27 \\ 
         \hline
         Discriminability & 0.12 & 0.08 & 0.19 & 0.03 & 0.16 & 0.40 & 0.22 & - \\
         \bottomrule[1pt]
    \end{tabular}
    \label{tab:llm_performance}
\end{table*}

\subsection{Redundancy and Diversity Analysis Across Benchmark Dimensions}
\label{sec:correlation_analysis}

\begin{figure}[t]
    \centering
    \includegraphics[width=0.95\textwidth]{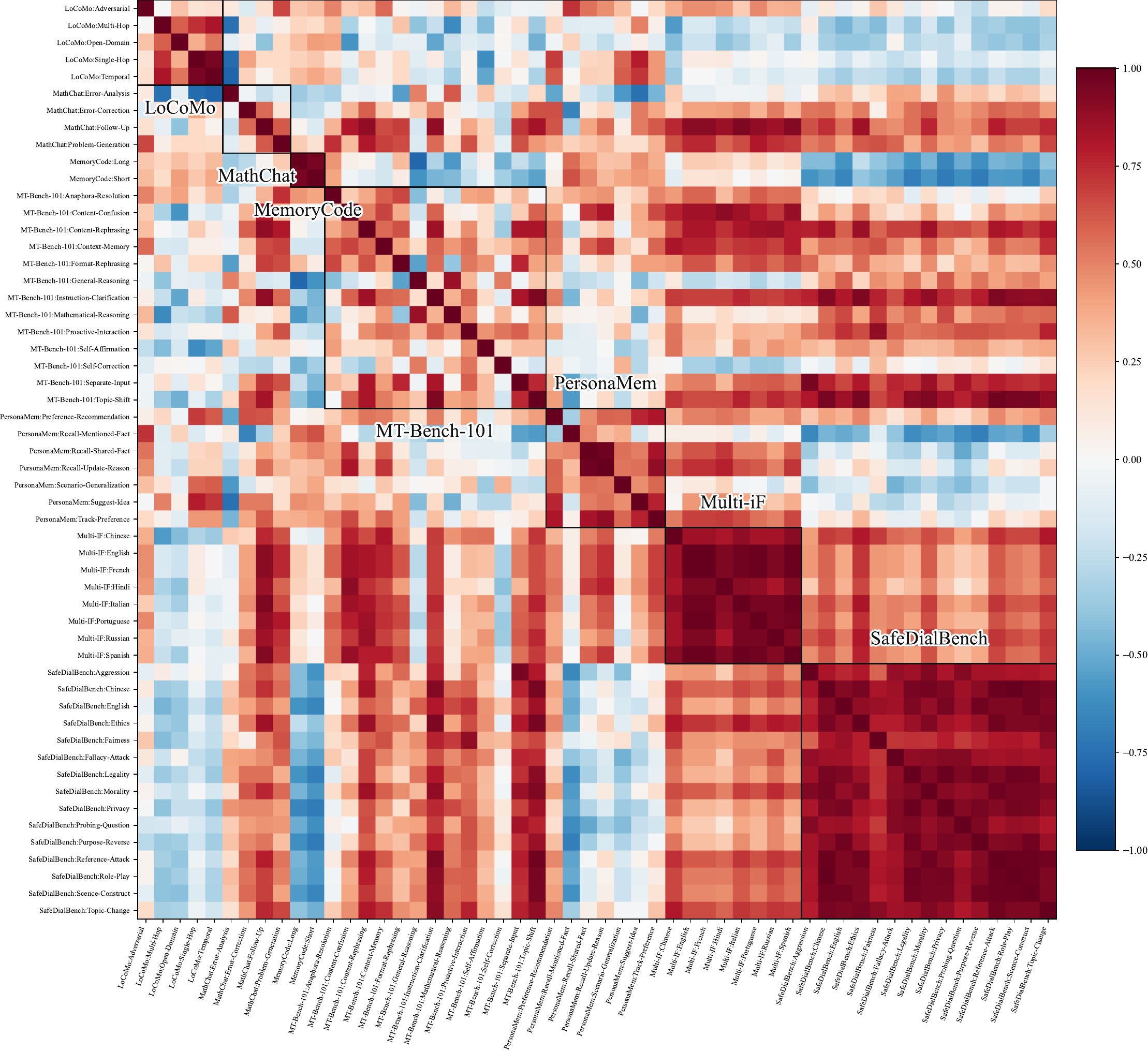}
    \caption{Spearman's rank correlation heatmap across all evaluation benchmark sub-categories.}
    \label{fig:correlation_heatmap}
\end{figure}

To systematically examine relationships between different evaluation capabilities, we conducted a correlation analysis of model performances across all sub-categories. Figure~\ref{fig:correlation_heatmap} visualizes Spearman's rank correlation ($\rho$), and further details are provided in Appendix~\ref{app:correlation_details}.  

\paragraph{Intra-Benchmark Analysis} Different benchmarks exhibit substantial internal differences. SafeDialBench shows high internal consistency, indicating that strong performance in one safety aspect generally reflects overall safety competence. Multi-IF also demonstrates alignment across languages, suggesting consistent multi-turn instruction-following capabilities. MemoryCode exhibits similar consistency between short- and long-history coding tasks. These observations also suggest potential \textbf{internal redundancy}, as highly correlated sub-dimensions produce nearly identical model rankings and could potentially be pruned without substantially affecting evaluation fidelity.
In contrast, MT-Bench-101 and MathChat demonstrate significant heterogeneity, containing multiple sub-dimension pairs that are uncorrelated or negatively correlated. MT-Bench-101 is particularly notable, with the strongest negative correlations occurring within the benchmark itself. For example, Format-Rephrasing and General-Reasoning exhibit a correlation of $-0.54$, indicating that its sub-tasks capture diverse and partially orthogonal interaction skills.

\paragraph{Inter-Benchmark Analysis} Beyond internal structures, off-diagonal regions in the heatmap reveal unexpected cross-benchmark relationships. Some seemingly unrelated datasets exhibit high positive correlations. For example, MT-Bench-101:Topic-Shift shows strong correlations with $\rho > 0.93$ across multiple safety dimensions in SafeDialBench, including Reference-Attack and Ethics. This suggests a \textbf{latent shared capability}: the contextual robustness required to handle abrupt conversational topic shifts aligns closely with the ability to navigate adversarial safety attacks. 
On the other hand, the strongest negative inter-dataset correlations highlight fundamentally \textbf{orthogonal evaluation axes}. MathChat:Error-Analysis, for instance, negatively correlates with LoCoMo several tasks with $\rho$ up to $-0.80$. Similarly, MemoryCode:Long exhibits a strong negative correlation with MT-Bench-101:General-Reasoning with $\rho = -0.78$. These findings indicate potential diverging optimization paths for LLMs, which should be considered during training.

\begin{figure}[t!]
    \centering
    \subfigure[Score vs. Output Length across Benchmarks]{
        \begin{minipage}[b]{0.69\textwidth}
        \includegraphics[width=1\textwidth]{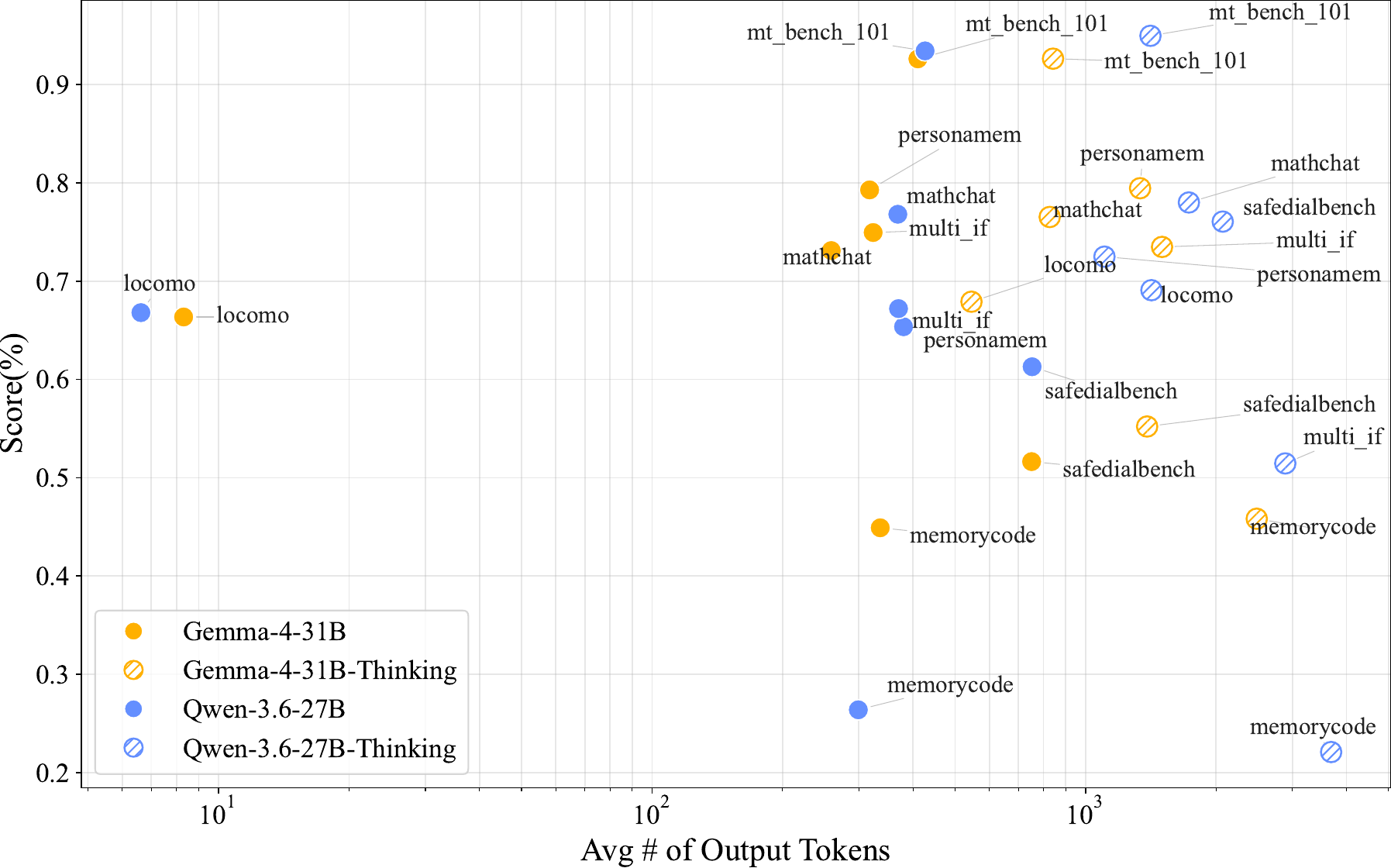}
        \end{minipage}
        \label{fig:thinking_score_vs_output}
    }
    \subfigure[Input-Output Token Breakdown]{
        \begin{minipage}[b]{0.27\textwidth}
        \includegraphics[width=1\textwidth]{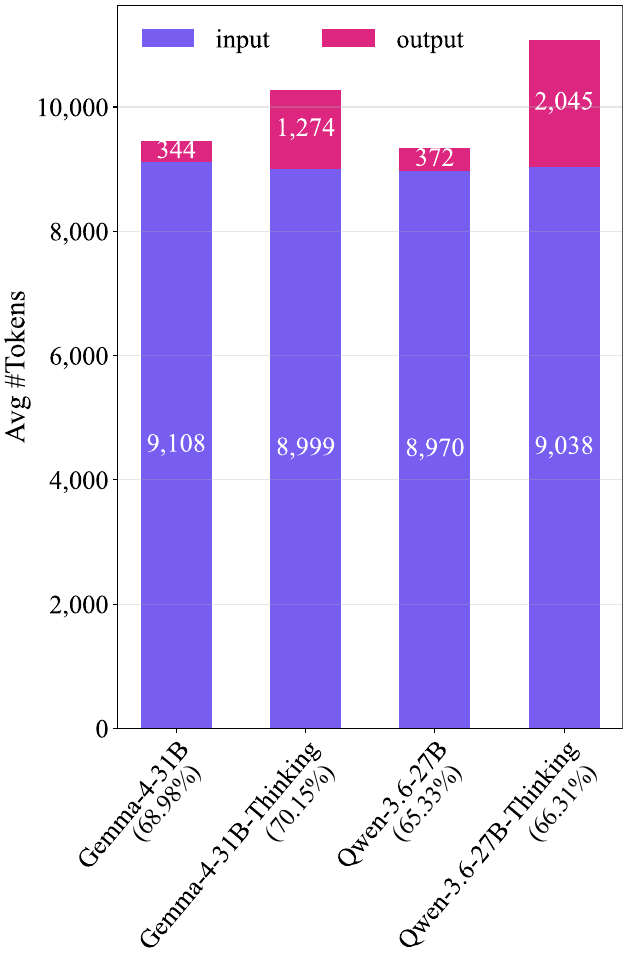}
        \end{minipage}
        \label{fig:thinking_token_breakdown}

    }
    \caption{Impact of thinking mode on model performance and token usage. In (b), values in parentheses denote the average benchmark scores of corresponding models.}
    \label{fig:thinking_mode_analysis}
\end{figure}

\subsection{Performance-Cost Trade-offs of Thinking Modes}
\label{sec:performance_cost}

To evaluate the performance-cost trade-off of integrating a thinking mode, we analyze the relationship between evaluation scores and token consumption across benchmarks, as shown in Figure~\ref{fig:thinking_mode_analysis} with more in Appendix~\ref{app:thinking_mode}. In Figure~\ref{fig:thinking_mode_analysis}(a), performance is plotted against average output length on a logarithmic scale for Gemma-4-31B and Qwen-3.6-27B under both modes. A pattern emerges: \textbf{thinking modes achieve higher scores on most benchmarks}, but at the cost of substantially longer generations.

Figure~\ref{fig:thinking_mode_analysis}(b) further quantifies this overhead by decomposing input and output token usage. Input lengths remain stable at around 9k tokens, while output lengths increase significantly. Standard models generate 344 and 372 tokens on average, whereas thinking mode expands this to $3.70\times$ for Gemma-4-31B and $5.5\times$ for Qwen-3.6-27B. Although thinking improves accuracy, it incurs more auto-regressive decoding steps, leading to higher latency and cost. Notably, \textbf{longer outputs do not necessarily yield better performance.} Despite Qwen-3.6-27B producing $1.61\times$ longer outputs than Gemma-4-31B, Gemma-4-31B still achieves a $3.84\%$ higher average score.

\subsection{Performance of Memory Agents}

Table~\ref{tab:agent_performance} reveals several findings on the effectiveness of memory agents. 

\begin{table*}[h!]
    \centering
    \small
    \caption{Performance of memory agents on normalized benchmarks.}
    \begin{tabular}{l|p{1.4cm}p{1.3cm}p{1.6cm}|p{1.4cm}p{1.3cm}p{1.6cm}}
        \toprule[1pt]
            & \multicolumn{3}{c|}{Gemma-4-31B} & \multicolumn{3}{c}{Qwen-3.6-27B} \\
            & PersonaMem & LoCoMo & MemoryCode & PersonaMem & LoCoMo & MemoryCode\\
        \midrule[1pt]
        Full Context & 79.29 & 69.05 & 44.91 & 65.37 & 66.80 & 26.39 \\
        \hdashline
        HippoRAG & $\underline{78.78}\downv{0.51}$ & $\textbf{47.98}\downv{21.07}$ & $\underline{23.35}\downv{21.56}$ & $\textbf{74.36}\up{8.99}$ & $\textbf{47.84}\downv{18.96}$ & $\underline{20.70}\downv{5.69}$ \\
        LightMem & $69.61\downv{9.68}$ & $34.63\downv{34.42}$ & $19.91\downv{25.00}$ & $53.31\downv{12.06}$ & 34.43$\downv{32.37}$ & $13.96\downv{12.43}$ \\
        RFMem & $\textbf{79.46}\up{0.26}$ & $43.59\downv{25.46}$ & $1.12\downv{43.79}$ & $64.69\downv{0.68}$ & $44.60\downv{22.20}$ & $8.20\downv{18.19}$ \\
        MemPalace & $78.43\downv{0.77}$ & $\underline{46.53}\downv{22.52}$ & $\textbf{25.33}\downv{19.58}$ & $\underline{67.23}\up{1.86}$ & $\underline{47.43}\downv{19.37}$ & $\textbf{23.08}\downv{3.31}$ \\
        \bottomrule[1pt]
    \end{tabular}    
    \label{tab:agent_performance}
\end{table*}

\textbf{No single agent consistently outperforms others across benchmarks.} HippoRAG shows gains on PersonaMem under Qwen-3.6-27B, while MemPalace performs relatively better on MemoryCode.
LightMem consistently degrades performance, suggesting that rewriting or compressing the original context may lose critical information. RFMem's iterative evidence expansion performs competitively on PersonaMem but fails dramatically on MemoryCode, implying limited generalization ability.

\textbf{Memory agents often fail to surpass the full-context baseline}. In most cases, directly feeding the complete context into the base model yields stronger or more stable performance, particularly on LoCoMo and MemoryCode. This suggests that retrieval or compression errors can offset the intended benefits of memory augmentation, and that information loss remains a central bottleneck. Overall, current memory agents remain unreliable substitutes for full-context reasoning.

\subsection{Effectiveness of Memory Retrieval}

Leveraging the explicit reference turns in LoCoMo, we evaluate the retrieval quality of different memory agents. To determine whether a retrieved memory snippet matches the ground-truth reference turn, we employ ROUGE-2~\citep{lin2004rouge} recall as the matching criterion, based on which retrieved hit@k and recall scores are computed. Results are in Table~\ref{tab:memory_retrieval_percent}. We use empirical thresholds of 1.0 for unmodified snippets and 0.2 for rewritten representations,

\begin{table}[h!]
\centering
\small
\caption{Memory retrieval performance (\%) across different agents.}
\begin{tabular}{l|ccc|ccc}
\toprule[1pt]
 & \multicolumn{3}{c|}{Gemma-4-31B} & \multicolumn{3}{c}{Qwen-3.6-27B} \\
Agent & hit@1/5/10 & Recall & Score & hit@1/5/10 & Recall & Score \\
\midrule[1pt]
HippoRAG & 13.60 / 29.59 / \textbf{38.21} & \textbf{37.89} & \textbf{47.98} & 10.86 / 26.35 / \textbf{35.61} & \textbf{35.33} & \textbf{47.84} \\
LightMem & 4.34 / 12.14 / 15.88 & 15.49 &  34.63 & 4.10 / 8.97 / 12.25 & 11.93 & 34.43 \\
RFMem & 7.80 / 21.05 / 28.77 & 28.35 & 43.59 & 7.80 / 21.05 / 28.77 & 28.35 & 44.60 \\
MemPalace & \textbf{25.39} / \textbf{31.37} / 34.94 & 34.65 & 46.53 & \textbf{26.32} / \textbf{31.70} / 34.94 & 34.65 & 47.43 \\

\bottomrule[1pt]
\end{tabular}
\label{tab:memory_retrieval_percent}
\end{table}

We observe\textbf{ a positive correlation between overall retrieval broadness}, specifically Hits@10 and Recall, \textbf{and the final response quality}. For instance, although MemPalace achieves notably higher Hits@1, HippoRAG secures superior Hits@10, which ultimately leads to marginally better downstream performance. Furthermore, cross-model evaluations reveal \textbf{differences in the robustness of different memory mechanisms}. Operating with minimal LLM reliance, RFMem and MemPalace exhibit exceptionally high cross-model robustness. Conversely, the retrieval accuracy of HippoRAG and LightMem fluctuates significantly, indicating a heavier reliance on model-specific capabilities during the critical memory construction phases. Moreover, \textbf{the ultimate generation performance relies on the LLM's intrinsic capacity} to filter noise and isolate actionable information from a cluster of retrieved memory pieces. For example, despite Qwen-3.6-27B achieving a lower HippoRAG Recall than Gemma-4-31B with 35.33\% versus 37.89\%, their final downstream scores remain similar.

\subsection{How Architectural Inductive Biases Shape Agent Capabilities?}

\begin{figure}[ht!]
    \centering
    \subfigure[LoCoMo]{
        \begin{minipage}[b]{0.4\textwidth}
        \includegraphics[width=1\textwidth]{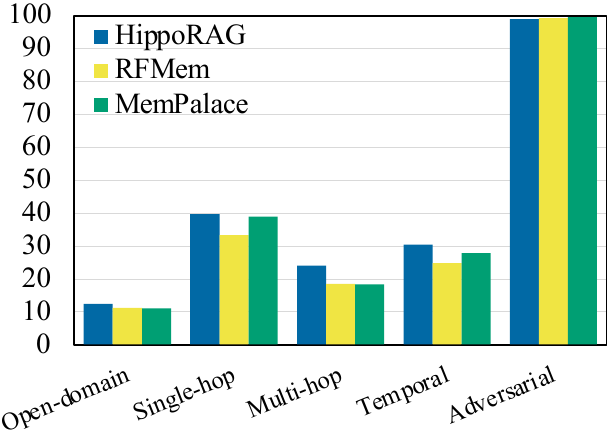}
        \end{minipage}
        \label{fig:finegrained_locomo}
    }
    \subfigure[PersonaMem]{
        \begin{minipage}[b]{0.5\textwidth}
        \includegraphics[width=1\textwidth]{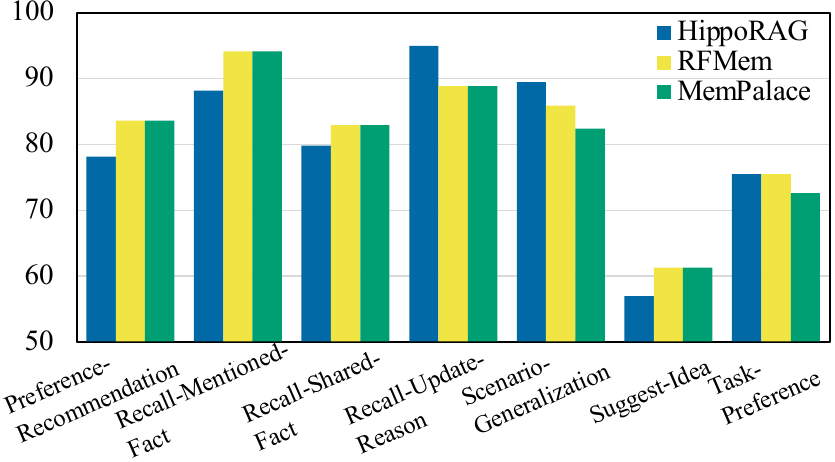}
        \end{minipage}
        \label{fig:finegrained_personamem}

    }
    \caption{Fine-grained performance of memory agents on LoCoMo and PersonaMem.}
    \label{fig:agent_finegrained_results}
\end{figure}

A fine-grained analysis of the benchmark results in Table~\ref{fig:agent_finegrained_results} reveals a profound correlation between the underlying architectural mechanisms of memory agents and their task-specific performance. 
Results show that current memory agents remain limited by their architectural designs. Specifically, {HippoRAG}'s integration of Knowledge Graphs and Personalized PageRank excels at complex logical routing, dominating LoCoMo's Multi-hop and Temporal dimensions where connecting disparate evidence is crucial. {MemPalace} employs a hybrid heuristic by fusing exact entity matching with LLM-based reranking. This precise entity-centric design makes it robust for exact fact retrieval, securing top scores in LoCoMo:Single-hop and PersonaMem:Recall-Mentioned-Fact. Meanwhile, {RFMem}'s familiarity-guided dual-path retrieval favors broad embedding-space semantic matching, yielding optimal performance in PersonaMem:Preference-Recommendation. Ultimately, these observations demonstrate that \textbf{while specific inductive biases allow agents to achieve state-of-the-art results in tailored scenarios, they act as a double-edged sword that may hinders broad generalizability}.

\section{Conclusion and Future Directions}
\label{sec:conclusion_future}

In this work, we introduced UniDial-EvalKit, an open-source framework for unified evaluation of multi-turn interactions. UDE provides a \textbf{standardized} dialogue schema for fair cross-benchmark comparisons, an \textbf{extensible} pipeline for integrating diverse LLMs and memory agents, and \textbf{comprehensive} evaluation coverage across memory, safety, mathematics, and etc. Looking forward, we plan to extend UDE with interactive user simulators, multimodal evaluation settings, and visualization modules for deeper behavioral analysis. We hope these open-source efforts will support reproducible research and advance conversational AI evaluation. Based on our empirical studies with UDE, we identify several future directions as follows:

\textbf{Evolution of Multi-Turn Evaluation Paradigms.} To accurately reflect the frontier of AI capabilities, evaluation benchmarks must evolve alongside foundation models. As LLMs continue to extend their context windows, benchmarks should scale in complexity to avoid rapid metric saturation. In addition, our analysis in Section~\ref{sec:correlation_analysis} shows that benchmark redundancy within and across datasets is highly prevalent and can distort evaluation results~\citep{redundancyprinciples,kim-etal-2025-benchmark}, calling for principled and unified data partitioning strategies targeting on multi-turn scenarios. Future datasets are also expected to provide more fine-grained annotations to enable transparent evidence tracking and error attribution. 

\textbf{Next-Generation Memory-Augmented Systems.} The design of memory agents also requires a paradigm shift. As long-context modeling abilities continue to improve, the relationship between internal parametric knowledge and external memory systems must be reconsidered. Within finite dialogue horizons, external memory may move beyond organizing valid information toward filtering noisy contexts and distilling long-term user preferences. Future memory mechanisms should further extend beyond semantic similarity retrieval and support more complex cognitive signals such as evolving user intentions and instructions. More adaptive memory architectures, where memory structures dynamically reorganize according to task requirements, may provide a promising direction for improving generalization across diverse scenarios.

\bibliography{custom}
\bibliographystyle{plainnat}

\appendix

\section{Unified Dialogue Schema}
\label{app:schema}
We present the detailed dialogue schema as follows:

\begin{verbatim}
Dialog = {
    "dialog_id": ...,
    "dialog_raw_info": {...},
    "dialog_labels": {...},
    "dialog_eval_config": {
        "use_reference_history": ...
    },
    "dialog_turns": [
        {
            "turn_id": ...,
            "role": ...,
            "content": ...,
            "reference": ...,
            "reference_document": ...,
            "eval_config": {
                "do_eval": ...,
                "metrics": [
                    {
                        "class_name": ...,
                        "args": {...}
                    },
                    ...
                ],
            },
            "turn_labels": {...}
        },
        ...
    ]
}
\end{verbatim}

\section{Details of Integrated Benchmarks}
\label{app:dataset}

UDE integrates over 10 benchmarks spanning diverse and multifaceted conversational abilities. Among them, seven representative benchmarks are summarized as follows:

\begin{itemize}[leftmargin=*]
    \item \textbf{MT-Bench-101}~\citep{bai2024mt} evaluates LLMs' \textbf{general} chat capabilities through a hierarchical ability taxonomy, highlighting perceptivity, adaptability, and interactivity in extended dialogues.
    
    \item \textbf{LoCoMo}~\citep{maharana2024evaluating} tests long-context \textbf{memory} with question-answering across single-hop, multi-hop, temporal, open-domain, and adversarial reasoning categories. 

    \item \textbf{PersonaMem}~\citep{jiang2025know} assesses \textbf{personalization} by evaluating LLMs' ability to extract, retain, and apply user-specific profiles across multiple multi-turn sessions and real-world tasks. The ``32k'' subset was adopted in current evaluation.

    \item \textbf{Multi-IF}~\citep{he2024multi} measures \textbf{instruction-following} proficiency in multi-turn, multilingual dialogues using a hybrid framework combining LLM and human annotations.  

    \item \textbf{SafeDialBench}~\citep{cao2026safedialbench} evaluates multi-turn \textbf{safety} under various jailbreak attacks using a two-tier hierarchical safety taxonomy covering six dimensions in over 4000 dialogues.  

    \item \textbf{MathChat}~\citep{liang2024mathchat} benchmarks multi-turn \textbf{mathematical} reasoning tasks such as follow-up QA, error correction, error analysis, and problem generation.  

    \item \textbf{MemoryCode}~\citep{rakotonirina-etal-2025-tools} tests \textbf{coding} capabilities in multi-turn programming interactions, including tracking instructions amid irrelevant information and iterative debugging.  
\end{itemize}

\section{Dataset Down-Sampling}
\label{sec:down_sampling}

Due to the large number of samples in the original datasets, including MathChat, MultiIF, and SafeDialBench, and the high evaluation cost, we performed label-stratified down-sampling to roughly 1000 samples per dataset. Table~\ref{tab:downsampling_results} presents model performance before and after down-sampling. The minimal differences suggest considerable redundancy in the datasets, highlighting the necessity of careful test set selection for benchmarking efficiency.

\begin{table}[h]
\centering
\small
\caption{Model performance before (Before) and after (After) down-sampling.}
\label{tab:downsampling_results}
\resizebox{\columnwidth}{!}{%
\begin{tabular}{l|cc|cc|cc|cc|cc}
\toprule[1pt]
Dataset & \multicolumn{2}{c|}{DeepSeek-V3.2} & \multicolumn{2}{c|}{Qwen3-Max} & \multicolumn{2}{c|}{Kimi-K2.5} & \multicolumn{2}{c|}{GLM-5} & \multicolumn{2}{c}{MiniMax-M2.5} \\
 & Before & After & Before & After & Before & After & Before & After & Before & After \\
\midrule[1pt]
MathChat & 77.87 & 77.79 & 77.43 & 77.31 & 61.41 & 60.47 & 67.65 & 69.15 & 69.58 & 70.54 \\
Multi-IF & 64.23 & 65.10 & 68.93 & 68.61 & 26.94 & 26.18 & 26.25 & 25.88 & 31.76 & 31.79 \\
SafeDialBench & 53.95 & 54.63 & 61.33 & 62.98 & 49.86 & 51.35 & 34.02 & 33.40 & 46.85 & 48.75 \\
\bottomrule[1pt]
\end{tabular}
}
\end{table}

\section{Detailed Correlation Analysis}
\label{app:correlation_details}

Table~\ref{tab:correlation_full} details the top-10 extreme correlation pairs, thoroughly categorized by intra-dataset and inter-dataset scopes. Positive correlations reveal several notable patterns across benchmark dimensions. In Multi-IF, different language variants exhibit near-identical rankings across models, suggesting that multilingual instruction-following performance is highly consistent across languages. Similarly, in SafeDialBench, multiple safety-related categories are strongly correlated, indicating that models which perform well on one type of safety scenario generally remain robust across other adversarial or unsafe settings. In contrast, negative correlations highlight orthogonal or even conflicting capability requirements. For example, multiple MT-Bench-101 dimensions related to rephrasing or reference resolution are negatively correlated with reasoning-oriented tasks, suggesting that strong performance on one interaction skill does not necessarily transfer to another.

At the inter-dataset level, several dimensions from different benchmarks also demonstrate unexpectedly high positive correlations, implying that ostensibly distinct benchmarks may still evaluate similar underlying behaviors. For instance, topic-shift handling in MT-Bench-101 is highly correlated with multiple adversarial settings in SafeDialBench. Conversely, strong negative correlations expose substantial capability divergence across domains, particularly between memory-intensive tasks from LoCoMo and PersonaMem and mathematical reasoning tasks from MathChat. These findings collectively suggest that capability overlap and entanglement are prevalent both within and across existing dialogue benchmarks, motivating future evaluation designs toward more systematically disentangled and diversified capability dimensions.

\begin{table}[htbp]
\centering
\footnotesize
\caption{Comprehensive list of the Top-10 extreme Spearman correlations ($\rho$) between benchmark sub-dimensions.} 
\label{tab:correlation_full}
\begin{tabular}{@{}c ll c@{}}
\toprule
\textbf{Rank} & \textbf{Dimension A} & \textbf{Dimension B} & \textbf{Spearman $\rho$} \\ \midrule

\multicolumn{4}{c}{\textbf{Part I: Intra-Dataset - Highest Positive Correlations}} \\ \midrule
1  & Multi-IF: English & Multi-IF: French & $ 1.000$ \\
2  & SafeDialBench: Chinese & SafeDialBench: Role-Play & $ 0.996$ \\
3  & Multi-IF: English & Multi-IF: Portuguese & $ 0.983$ \\
4  & Multi-IF: English & Multi-IF: Spanish & $ 0.983$ \\
5  & Multi-IF: French & Multi-IF: Portuguese & $ 0.983$ \\
6  & Multi-IF: French & Multi-IF: Spanish & $ 0.983$ \\
7  & Multi-IF: Italian & Multi-IF: Spanish & $ 0.983$ \\
8  & SafeDialBench: Chinese & SafeDialBench: Reference-Attack & $ 0.983$ \\
9  & SafeDialBench: Legality & SafeDialBench: Privacy & $ 0.983$ \\
10 & SafeDialBench: Reference-Attack & SafeDialBench: Scence-Construct & $ 0.983$ \\ \midrule

\multicolumn{4}{c}{\textbf{Part II: Intra-Dataset - Strongest Negative Correlations}} \\ \midrule
1  & MT-Bench-101: Format-Rephrasing & MT-Bench-101: General-Reasoning & $-0.536$ \\
2  & MT-Bench-101: Anaphora-Resolution & MT-Bench-101: General-Reasoning & $-0.427$ \\
3  & MT-Bench-101: Format-Rephrasing & MT-Bench-101: Mathematical-Reasoning & $-0.387$ \\
4  & MT-Bench-101: Content-Rephrasing & MT-Bench-101: Self-Correction & $-0.358$ \\
5  & PersonaMem: Preference-Recommendation & PersonaMem: Recall-Mentioned-Fact & $-0.329$ \\
6  & MT-Bench-101: Content-Confusion & MT-Bench-101: Self-Correction & $-0.269$ \\
7  & MT-Bench-101: Anaphora-Resolution & MT-Bench-101: Mathematical-Reasoning & $-0.235$ \\
8  & MathChat: Error-Analysis & MathChat: Follow-Up & $-0.233$ \\
9  & MT-Bench-101: Content-Confusion & MT-Bench-101: General-Reasoning & $-0.210$ \\
10 & MT-Bench-101: Self-Correction & MT-Bench-101: Topic-Shift & $-0.192$ \\ \midrule\midrule

\multicolumn{4}{c}{\textbf{Part III: Inter-Dataset - Highest Positive Correlations}} \\ \midrule
1  & MT-Bench-101: Topic-Shift & SafeDialBench: Reference-Attack & $ 0.967$ \\
2  & MT-Bench-101: Separate-Input & SafeDialBench: Aggression & $ 0.950$ \\
3  & MT-Bench-101: Topic-Shift & SafeDialBench: Chinese & $ 0.950$ \\
4  & MT-Bench-101: Topic-Shift & SafeDialBench: Ethics & $ 0.950$ \\
5  & MT-Bench-101: Topic-Shift & SafeDialBench: Scence-Construct & $ 0.950$ \\
6  & MT-Bench-101: Topic-Shift & SafeDialBench: Role-Play & $ 0.946$ \\
7  & MT-Bench-101: Topic-Shift & SafeDialBench: Morality & $ 0.937$ \\
8  & MT-Bench-101: Instruction-Clarification & SafeDialBench: Ethics & $ 0.933$ \\
9  & MT-Bench-101: Instruction-Clarification & SafeDialBench: Reference-Attack & $ 0.933$ \\
10 & MathChat: Follow-Up & Multi-IF: Italian & $ 0.933$ \\ \midrule

\multicolumn{4}{c}{\textbf{Part IV: Inter-Dataset - Strongest Negative Correlations}} \\ \midrule
1  & LoCoMo: Temporal & MathChat: Error-Analysis & $-0.800$ \\
2  & MT-Bench-101: General-Reasoning & MemoryCode: Long & $-0.783$ \\
3  & LoCoMo: Single-Hop & MathChat: Error-Analysis & $-0.767$ \\
4  & LoCoMo: Multi-Hop & MathChat: Error-Analysis & $-0.750$ \\
5  & MathChat: Error-Analysis & PersonaMem: Suggest-Idea & $-0.750$ \\
6  & MemoryCode: Short & SafeDialBench: Privacy & $-0.700$ \\
7  & MemoryCode: Short & SafeDialBench: Purpose-Reverse & $-0.700$ \\
8  & MemoryCode: Short & SafeDialBench: Legality & $-0.683$ \\
9  & PersonaMem: Recall-Mentioned-Fact & SafeDialBench: Purpose-Reverse & $-0.682$ \\
10 & MathChat: Error-Correction & PersonaMem: Recall-Mentioned-Fact & $-0.665$ \\ \bottomrule
\end{tabular}
\end{table}

\section{Impact of Thinking Modes}
\label{app:thinking_mode}

To investigate the efficacy of extended inference-time reasoning in continuous interactions, we evaluate the ``thinking modes'' of Gemma-4-31B and Qwen-3.6-27B. To accommodate the verbose chain-of-thought traces inherent to these modes, we relax the generation constraint to 4,096 tokens per turn, while the non-thinking baselines adhere to the standard evaluation setup detailed in Section~\ref{sec:evaluation_configuration}. The comparative results, summarized in Table~\ref{tab:llm_w_wo_thinking}, reveal highly consistent capability shifts across both foundational models.

Activating the thinking mode yields a marginal but consistent improvement in the overall average scores for both models. A fine-grained analysis reveals that this extended reasoning paradigm universally benefits domains requiring logical deduction and multi-step safety alignments. Specifically, both models demonstrate marked performance gains in \texttt{MathChat}, \texttt{LoCoMo}, and \texttt{SafeDialBench}. Conversely, the thinking mode introduces vulnerability in strict instruction-following scenarios. Notably, performance on the \texttt{Multi-IF} benchmark drops for both models with Qwen experiencing a drastic plummet from 67.21 to 51.46. Ultimately, these findings underscore that while thinking modes elevate ceilings, they do not serve as a universal panacea in multi-turn dialogues, highlighting a critical trade-off between complex reasoning capacity and strict output controllability.

\begin{table}[h!]
    \centering
    \small
    \caption{Performance comparisons with and without thinking modes. Bold values indicate the superior setup for each specific model.}
    \resizebox{\columnwidth}{!}{%
    \begin{tabular}{l|ccccccc|c}
    \toprule[1pt]
        \textbf{Model Setup} & \textbf{LoCoMo} & \makecell{\textbf{Math}\\\textbf{Chat}} & \makecell{\textbf{Memory}\\\textbf{Code}} & \makecell{\textbf{MT-Bench}\\\textbf{-101}} & \makecell{\textbf{Persona}\\\textbf{Mem}} & \textbf{Multi-IF} & \makecell{\textbf{SafeDial}\\\textbf{Bench}} & \textbf{Avg} \\
     \midrule[1pt]

        Gemma-4-31B & 66.36 & 73.11 & 44.91 & 92.62 & 79.29 & \textbf{74.95} & 51.64 & 68.98 \\
       \textit{w.} thinking & \textbf{67.91} & \textbf{76.53} & \textbf{45.83} & \textbf{92.64} & \textbf{79.46} & 73.48 & \textbf{55.21} & \textbf{70.15}\\
       \hline

       Qwen-3.6-27B & 66.80 & 76.81 & \textbf{26.39} & 93.43 & 65.37 & \textbf{67.21} & 61.29 &  65.33 \\
       \textit{w.} thinking & \textbf{69.08} & \textbf{78.01} & 22.09 & \textbf{94.96} & \textbf{72.50} & 51.46 & \textbf{76.06} & \textbf{66.31} \\
    \bottomrule[1pt]
    \end{tabular}
    }
    \label{tab:llm_w_wo_thinking}
\end{table}

\section{Comparisons on Various Score Aggregation Settings}
\label{app:aggregation_choice}

\begin{figure}[h]
    \centering
    \includegraphics[width=0.7\linewidth]{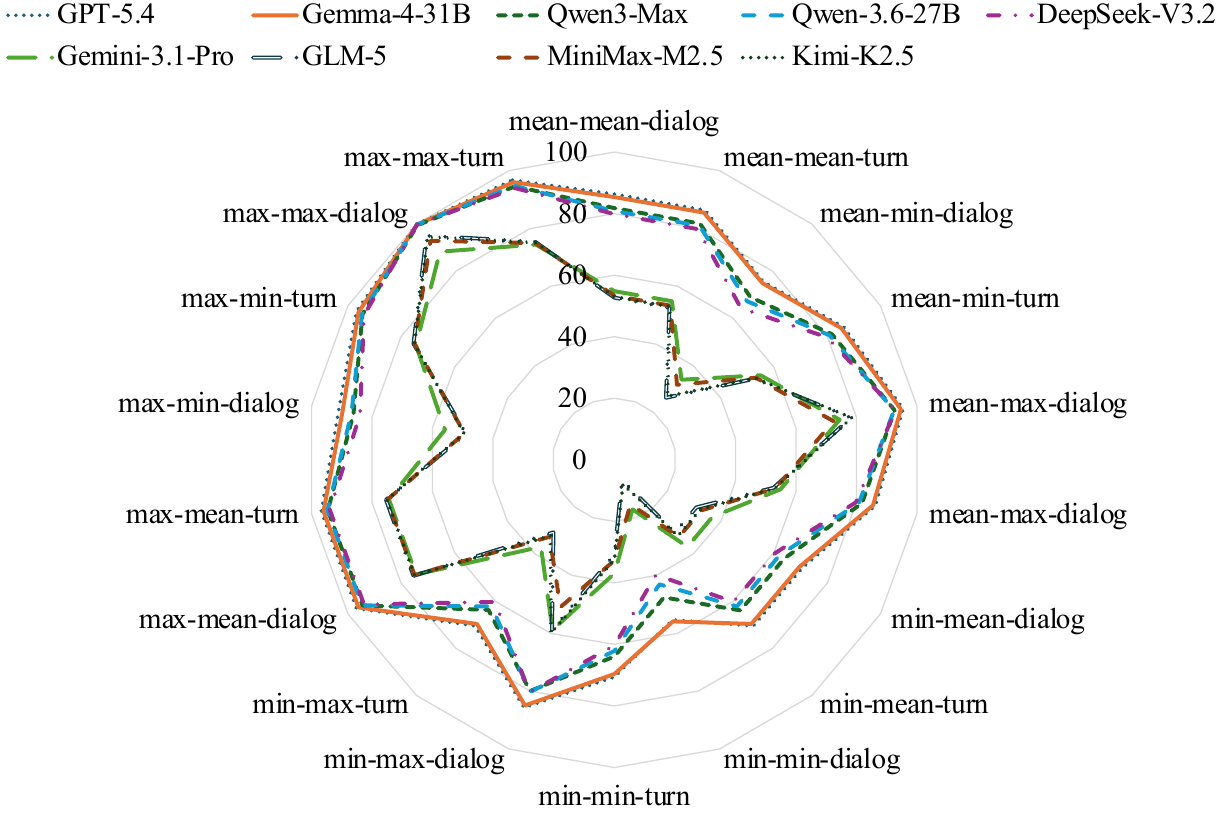}
    \caption{Performance across Aggregation Settings}
    \label{fig:aggregation_choice}
\end{figure}

We further analyze the impact of different score aggregation strategies using Multi-IF as a representative example. Figure~\ref{fig:aggregation_choice} presents the results under various aggregation settings.

Overall, model rankings remain relatively stable across different aggregation strategies, suggesting that the comparative conclusions between models are generally robust. Nevertheless, local ranking fluctuations still exist under certain settings. For example, under the stricter \texttt{min-min-dialog} setting, Qwen-3-Max drops from 81.73\% under \texttt{mean-mean-dialog} to 47.50\%, while GLM-5 further decreases from 52.63\% to only 7.13\%. In contrast, GPT-5.4 and Gemma-4-31B remain relatively more stable, achieving 55.44\% and 55.83\% respectively under the same setting. These results suggest that stronger models maintain more consistent instruction-following performance across dialogue turns, whereas weaker or middle-tier models are more vulnerable to occasional failures that are amplified by minimum-based aggregation.

More importantly, the absolute scores vary substantially across aggregation choices. For instance, GPT-5.4 ranges from 55.44\% under the \texttt{min-min-dialog} setting to 100.00\% under \texttt{max-max-dialog}, while GLM-5 ranges from only 7.13\% to 95.05\% under the same configurations. These dramatic differences indicate that \textbf{aggregation strategy itself can significantly affect the reported benchmark difficulty and model performance}. Consequently, inconsistent or insufficiently specified aggregation settings may easily introduce unfair comparisons across different works. This observation further motivates the need for standardized and transparent evaluation protocols in multi-turn dialogue benchmarks.

\section{Impact of Output Normalization on Memory Agent Evaluation}
\label{app:input_output}

Prior to our comprehensive benchmarking, we conduct a pilot experiment to investigate the potential confounding effects of agent-specific prompt engineering. Existing memory agents often encapsulate highly specific, internal prompt templates to guide the LLM's final response generation after the memory retrieval phase. To rigorously isolate the impact of these built-in prompts, we evaluate HippoRAG using both its encapsulated \textit{original} prompt and a \textit{normalized} setup.

As detailed in Table~\ref{tab:hipporag_norm}, the discrepancies between the two setups are conspicuous. Most notably, Qwen-3.6-27B experiences a catastrophic failure using the original prompt on PersonaMem with 22.58\% but surges to 74.36 once the output format is normalized. Similarly, both models witness a massive absolute increase of approximately 20 points on LoCoMo. These empirical shifts powerfully demonstrate that unstandardized generation process can severely skew evaluation results. Driven by this finding, we enforce a strict standardization of the input-output generation process across all agents in our UDE framework. This unified pipeline guarantees that our evaluations fairly and exclusively benchmark the underlying \textit{memory retrieval and construction mechanisms}, free from prompt-induced artifacts.

\begin{table}[htbp]
    \centering
    \small
    \caption{Pilot evaluation of HippoRAG comparing its encapsulated original prompt against a normalized prompt for final answer generation.} 
    \begin{tabular}{l|cc|cc|cc}
    \toprule[1pt]
         \multirow{2}{*}{\textbf{Model}} & \multicolumn{2}{c|}{\textbf{PersonaMem}} & \multicolumn{2}{c|}{\textbf{LoCoMo}} & \multicolumn{2}{c}{\textbf{MemoryCode}} \\
         \cmidrule(lr){2-3} \cmidrule(lr){4-5} \cmidrule(lr){6-7}
         & Original & Normalized & Original & Normalized & Original & Normalized \\
    \midrule[1pt]
    Gemma-4-31B & 77.42 & \textbf{78.78} & 29.17 & \textbf{47.98} & 22.62 & \textbf{23.35} \\
    Qwen-3.6-27B & 22.58 & \textbf{74.36} & 24.99 & \textbf{47.84} & 19.64 & \textbf{20.70} \\
    \bottomrule[1pt]
    \end{tabular}
    \label{tab:hipporag_norm}
\end{table}

\section{Bootstrap Confidence Intervals of the Main Results}
\label{app:confidence_interval}

To evaluate the statistical stability of the reported results, we estimate confidence intervals using percentile bootstrap resampling at the dialogue level. Specifically, we repeatedly resample dialogue sessions with replacement and recompute the final benchmark scores under the same aggregation configuration as the main experiments. All reported intervals correspond to 95\% confidence intervals with 1,000 bootstrap samples.

Tables~\ref{tab:llm_performance_ci} and~\ref{tab:agent_performance_ci} present the resulting confidence intervals for LLMs and memory agents, respectively. Overall, most benchmarks exhibit relatively narrow confidence ranges, suggesting that the main conclusions and model rankings are statistically stable under dialogue-level sampling variations.

\begin{table*}[ht!]
    \centering
    \small
    \caption{95\% percentile bootstrap confidence intervals (\%) of LLM performance on benchmarks.} 
    \resizebox{\columnwidth}{!}{%
    \begin{tabular}{l|ccccccc}
        \toprule[1pt]
         & LoCoMo & \makecell{Math\\Chat} & \makecell{Memory\\Code} & \makecell{MT-Bench\\-101} & \makecell{Persona\\Mem} & Multi-IF & \makecell{SafeDial\\Bench} \\
         \midrule[1pt]
         GPT-5.4 &  [59.83, 62.28] & [79.02, 82.58] & [40.36, 44.04] & [92.61, 93.89] & [67.30, 72.78] & [73.78, 77.11] & [64.73, 69.34] \\
         Gemma-4-31B & [65.23, 67.57] & [71.10, 75.02] & [42.89, 46.86] & [91.95, 93.30] & [76.92, 81.67] & [73.36, 76.54] & [49.40, 53.93] \\
         Qwen3-Max & [60.84, 63.32] & [76.56, 78.24] & [32.34, 35.77] & [92.96, 94.30] & [62.33, 67.98] & [68.08, 69.71] & [60.26, 62.46] \\
         Qwen-3.6-27B & [65.54, 68.04] & [74.82, 78.91] & [24.71, 28.23] & [92.81, 94.15] & [62.40, 67.99] & [65.67, 68.92] & [59.31, 63.50]\\
         DeepSeek-V3.2 & [59.89, 60.53] & [77.02, 78.66] &  [24.18, 27.62] & [90.38, 92.04] & [61.64, 67.40] & [63.39, 64.99] & [52.80, 54.98] \\
         Gemini-3.1-Pro & [67.82, 70.21] & [73.32, 77.09] & [37.73, 41.65] & [88.51, 90.16] & [76.88, 81.67] & [32.22, 35.75] & [31.61, 36.07] \\
         GLM-5 & [68.07, 70.37] & [66.69, 68.60] & [42.77, 46.81] & [84.54, 86.50] & [60.11, 66.21] & [25.51, 27.01] & [32.87, 35.15] \\
         MiniMax-M2.5 & [64.90, 67.28] & [68.61, 70.48] & [32.28, 35.97] & [89.13, 90.83] & [39.63, 45.87] & [30.98, 32.66] & [45.62, 48.04] \\
         Kimi-K2.5 & [41.50, 44.34] & [60.45, 62.33] & [30.00, 33.72] &  [89.95, 91.58] & [52.17, 58.50] & [26.18, 27.65] & [48.59, 50.95] \\ 
         \bottomrule[1pt]
    \end{tabular}
    }
    \label{tab:llm_performance_ci}
\end{table*}

\begin{table*}[ht!]
    \centering
    \small
    \caption{95\% percentile bootstrap confidence intervals (\%) of memory agent performance on normalized benchmarks. } 
    \resizebox{\columnwidth}{!}{%
    \begin{tabular}{l|ccc|ccc}
        \toprule[1pt]
            & \multicolumn{3}{c}{Gemma-4-31B} & \multicolumn{3}{c}{Qwen-3.6-27B} \\
            & PersonaMem & LoCoMo & MemoryCode & PersonaMem & LoCoMo & MemoryCode\\
        \midrule[1pt]
        HippoRAG & [76.33, 81.43] & [46.39, 49.48] & [21,74, 25.00] & [71.43, 76.97] & [46.21, 49.31] & [19.20, 22.23] \\
        LightMem & [66.84, 72.60] & [33.19. 36.02] & [18.35, 21.41] & [50.38, 56.29] & [32.91, 35.89] & [12.63, 15.30] \\
        RFMem & [77.07, 82.04] & [42.00, 45.13] & [0.72, 1.48] & [61.60, 67.64] & [43.00, 46.08] & [7.15, 9.31] \\
        MemPalace & [75.84, 80.87] & [44.95, 48.02] & [23.79, 26.97] & [64.23, 70.28] & [45.83, 48.94] & [21.44, 24.77] \\
        \bottomrule[1pt]
    \end{tabular} 
    }
    \label{tab:agent_performance_ci}
\end{table*}

\end{document}